\documentclass[conference]{IEEEtran}
\IEEEoverridecommandlockouts
\usepackage{cite}
\usepackage{amsmath,amssymb,amsfonts}
\usepackage{algorithmic}
\usepackage{graphicx}
\usepackage{textcomp}
\usepackage{xcolor}
\def\BibTeX{{\rm B\kern-.05em{\sc i\kern-.025em b}\kern-.08em
    T\kern-.1667em\lower.7ex\hbox{E}\kern-.125emX}}

\usepackage{times}
\usepackage{latexsym}

\usepackage[T1]{fontenc}

\usepackage[utf8]{inputenc}

\usepackage{microtype}

\usepackage{inconsolata}

\usepackage{graphicx}

\usepackage{algorithm}
\usepackage{algorithmic}

\usepackage{amsmath}
\usepackage{booktabs}
\usepackage{multirow}
\usepackage{colortbl}
\usepackage{caption}
\usepackage{hyperref}

\begin{document}

\title{LLM-Barber: Block-Aware Rebuilder for Sparsity Mask in One-Shot for Large Language Models\thanks{*These authors contributed equally to this work.}
}

\author{
\IEEEauthorblockN{
Yupeng Su\textsuperscript{*1}, 
Ziyi Guan\textsuperscript{*2}, 
Xiaoqun Liu\textsuperscript{1}, 
Tianlai Jin\textsuperscript{1}, 
Dongkuan Wu\textsuperscript{1}, 
Zhengfei Chen\textsuperscript{1},\\
Graziano Chesi\textsuperscript{2}, 
Ngai Wong\textsuperscript{2}, 
Hao Yu\textsuperscript{1}\textsuperscript{\dag}
}
\IEEEauthorblockA{
\textsuperscript{1}School of Microelectronics, Southern University of Science and Technology, Shenzhen, China\\
\textsuperscript{2}Department of Electrical and Electronic Engineering, University of Hong Kong, Hong Kong, China\\
\textsuperscript{\dag}Corresponding author: yuh3@sustech.edu.cn
}
}

\maketitle

\begin{abstract}
Large language models (LLMs) have seen substantial growth, necessitating efficient model pruning techniques. Existing post-training pruning methods primarily measure weight importance in converged dense models, often overlooking changes in weight significance during the pruning process, leading to performance degradation. 
To address this issue, we present LLM-\textbf{Barber} (\textbf{B}lock-\textbf{A}ware \textbf{R}e\textbf{b}uild\textbf{er} for Sparsity Mask in One-Shot), a novel one-shot pruning framework that rebuilds the sparsity mask of pruned models without any retraining or weight reconstruction. LLM-Barber incorporates block-aware error optimization across Self-Attention and MLP blocks, facilitating global performance optimization. 
We are the first to employ the product of weights and gradients as a pruning metric in the context of LLM post-training pruning. This enables accurate identification of weight importance in massive models and significantly reduces computational complexity compared to methods using second-order information. 
Our experiments show that LLM-Barber efficiently prunes models from LLaMA and OPT families (7B to 13B) on a single A100 GPU in just 30 minutes, achieving state-of-the-art results in both perplexity and zero-shot performance across various language benchmarks.
\end{abstract}


\section{Introduction}

LLMs have become foundational in artificial intelligence due to their impressive performance on various tasks. However, the increasing size and complexity of models, such as GPT-175B \cite{brown2020language} with 175 billion parameters, pose significant challenges related to extensive computational and storage demands.
Consequently, efficient model compression strategies are crucial for enabling the practical deployment of these powerful models.

\begin{figure}[t]
  \centering
  \includegraphics[width=\linewidth]{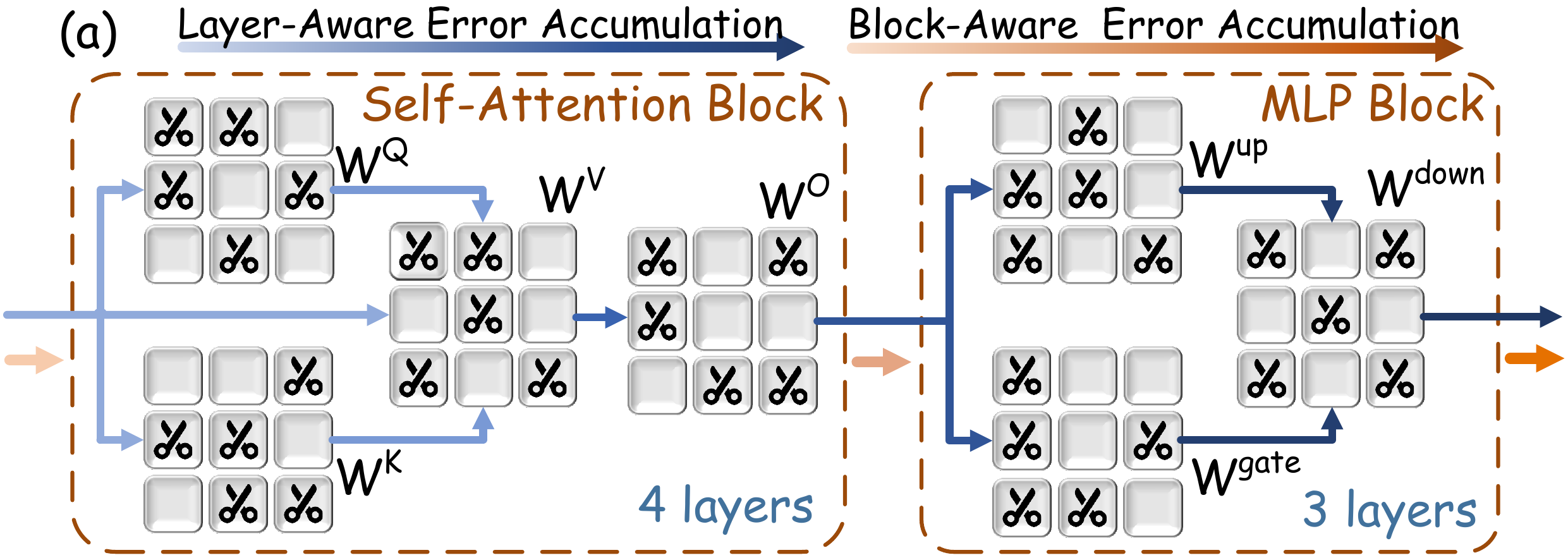}
  \includegraphics[width=1.0\linewidth]{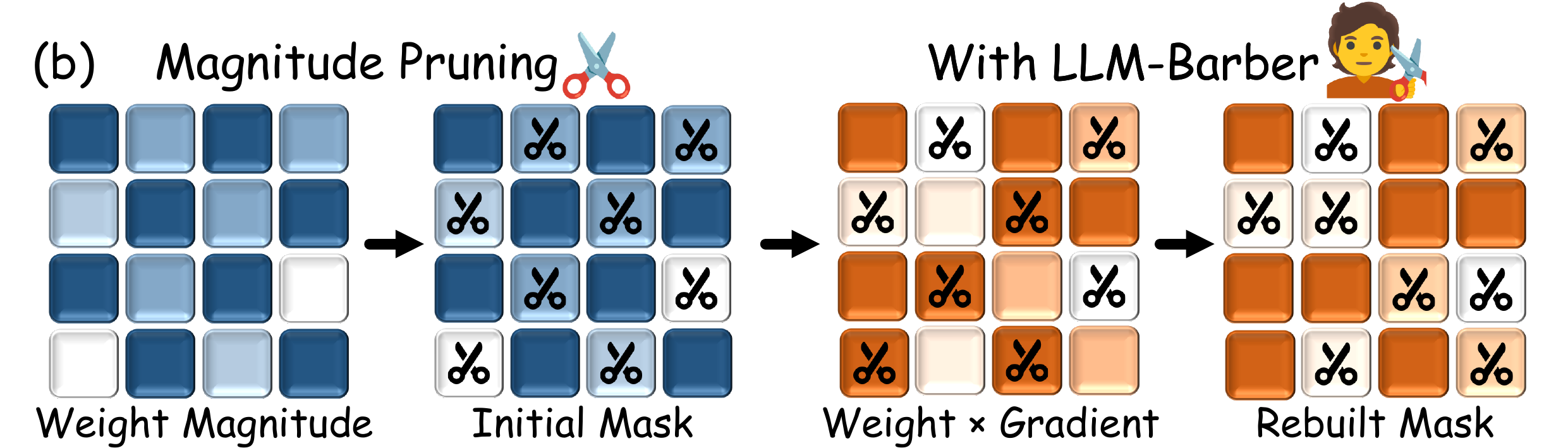}
  \captionsetup{justification=raggedright}
  \caption{The benefits of integrating LLM-Barber into the pruning process: (a) Transition from the layer-aware to block-aware error accumulation to achieve an optimized global solution. (b) Rebuilding sparsity mask using a novel pruning metric based on weights multiplied by gradients.}
  \label{fig:Overview}
\end{figure}

Current pruning methods face two major challenges. First, traditional layer-aware pruning methods focus on individual layers, neglecting inter-layer dependencies, which increases error accumulation (blue arrows in Figure \ref{fig:Overview}(a)). In contrast, block-aware pruning considers groups of layers, effectively reducing this error (orange arrows).
Second, as shown in Figure \ref{fig:Overview}(b), conventional methods typically build the pruning mask only once, ignoring the changes of weight significance in post-pruning stage. 
This oversight can lead to improper identification of salient weights and subsequent performance degradation.

To address these limitations, we propose \textbf{LLM-Barber}, a novel and straightforward approach rebuild sparsity mask of pruned networks without requiring for retraining or weight reconstruction.

\begin{figure*}[ht]
  \centering
  \includegraphics[width=\textwidth]{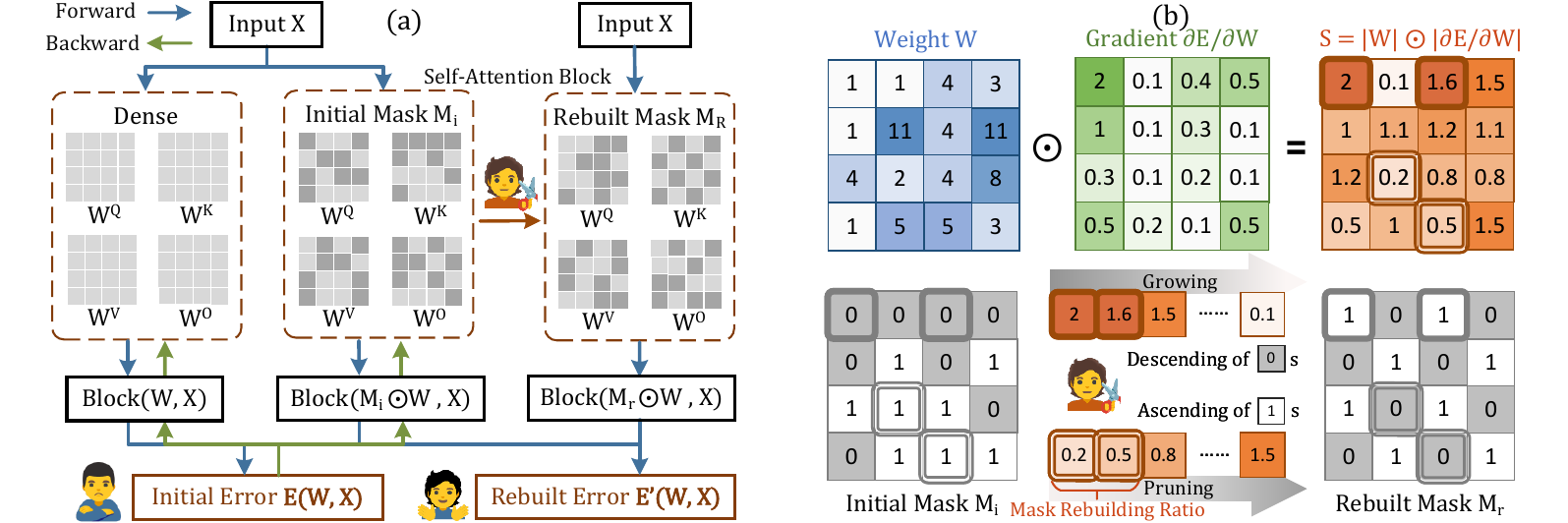}
  \captionsetup{justification=raggedright}
  \caption{The workflow of LLM-Barber. (a) illustrates the process of block-aware reconstruction error and gradient calculation for each linear weight. (b) shows pruning metric computation and  sparsity mask rebuilding.}
  \label{fig:Pipeline}
\end{figure*}
Firstly, unlike layer-aware methods that prioritize local optimization and are prone to error accumulation (illustrated by the blue arrows in Figure \ref{fig:Overview}(a)), LLM-Barber integrates pruning across Self-Attention and MLP blocks. This approach mitigates error accumulation, as evidenced by the lighter orange arrows, promoting more global optimization. 
Secondly, LLM-Barber identifies weights that, although initially non-salient without a sparsity mask, gain significance in post-pruning.
As shown in Figure \ref{fig:Overview}(b), varying color shades represent the relative importance scores of different weights. LLM-Barber accurately rebuilds masks for high-score weights (deeply shaded) while pruning newly identified low-score weights (lightly shaded).
thirdly, LLM-Barber employs the product of weights and gradients as pruning metric. While this metric has been applied in other contexts \cite{molchanov2019pruning, lee2019snip}, we are the first to leverage it in the context of post-training compression for LLMs. This enables precise rebuilding of the sparsity mask and significantly reduces computational complexity compared to methods \cite{frantar2023sparsegpt} using second-order information.
Finally, our one-shot approach efficiently rebuilds sparsity masks for pruned models, achieving comparable accuracy to iterative fine-tuning. 

LLM-Barber demonstrates its effectiveness through experiments on the LLaMA and OPT families, consistently outperforms existing post-training techniques in both perplexity and zero-shot tasks.
To sum up, the key contributions are fourfold:
\begin{itemize}
\item \textbf{Block-Aware Global Optimization:} We are among the first to introduce a block-aware reconstruction problem that integrates sparsity across the Self-Attention and MLP blocks, achieving global optimization in pruning.
\item \textbf{Rebuilding Sensitive Regions:} We identify non-salient weights that restore significance in post-pruning stage and rebuild sparsity mask to retain these weights, while simultaneously adjusting the mask to prune newly non-salient weights. This targeted rebuilding enhances overall model performance by optimally reallocating sparsity.
\item \textbf{Innovative Pruning Metric:} We are the first to apply the product of weights and gradients as a pruning metric within the context of LLM post-training pruning, leveraging first-order Taylor series for importance evaluation to reduce computational complexity compared to second-order Hessian-based approaches.
\item \textbf{Versatility and Efficiency:} LLM-Barber demonstrates its efficiency and effectiveness across various pruning techniques, consistently achieving state-of-the-art performance in perplexity and zero-shot tasks, thereby establishing new yardsticks in LLM post-training pruning.
\end{itemize}

\section{Related Work}
\subsection{LLMs Pruning and Sparsity} 
Network pruning reduces deep neural networks by eliminating unnecessary weights, with methods categorized into parameter-efficient fine-tuning (PEFT) and post-training approaches. 
PEFT begins with an initialized sparse network and refines it through iterative processes \cite{liu2019metapruning}. LoRA \cite{hulora} adapts pre-trained models to specific tasks by injecting trainable rank decomposition matrices. Dynamic Spase No Training \cite{zhang2024dynamic} prunes and grows weights to minimize reconstruction error. However, fine-tuning often requires ample data and can degrade performance.
Post-training, on the other hand,
removes weights from a pre-trained model. SparseGPT \cite{frantar2023sparsegpt} uses Hessian-based metrics and subsequent residual weight updates, Wanda \cite{sun2023simple} introduces a first-order metric using weight-activation products.
LLM-Barber employs a block-aware reconstruction strategy and rebuilds masks with a novel pruning metric.

\subsection{Model Compression Strategy}
Compression is key to reducing the memory and computational demands of model. 
Layer-aware strategies, originating from Optimal Brain Damage \cite{lecun1989optimal} and Optimal Brain Surgeon \cite{hassibi1993optimal}, have been enhanced by recent works like GPTQ \cite{frantar-gptq} using second-order information. 
However, block-aware strategies typically yield better accuracy recovery. For instance, APTQ \cite{guan2024aptq} applies global quantization to attention mechanisms, BESA \cite{xu2024besa} uses block-wise sparsity allocation. Our method leverages block-aware pruning to optimize global performance across blocks, effectively balancing efficiency and accuracy.

\section{LLM-Barber}

\subsection{Preliminaries}
LLM pruning removes weights from dense networks to minimize output discrepancies, which is computationally intensive cross large-scale models, leading to address a layer-aware reconstruction problem \cite{hassibi1993optimal}. This section reviews and reanalyses layer-aware reconstruction error and Taylor expansion at dense networks.

\textbf{Layer-aware Reconstruction Error.} For linear projection layer weight {$\mathbf{W}$} of shape ${(C_{\text{out}},C_{\text{in}})}$, where $C_{\text{out}}$, $C_{\text{in}}$ indicates the output and input channels. With $N$ calibration samples and sequence length $L$, the input activation is denoted as $\mathbf{X}$ with the shape of $(C_{\text{in}}, N\times L)$. 
Layer-aware reconstruction error $\mathbf{E}$ is defined as the $\ell_2$ norm difference between output of dense and sparse layers:
\begin{equation}
\mathbf{E}=||\mathbf{WX}-\mathbf{\widehat WX}||_2^2,
\label{eq:layererror1}
\end{equation}
where $\mathbf{\widehat{W}}$ is the element-wise product of $\mathbf{W}$ and a binary sparsity mask $\mathbf{M}(i,j) \in \{0,1\}$ of shape $(C_{\text{out}},C_{\text{in}})$, in context with mask selection and without weight reconstruction:
\begin{equation}
\mathbf{E}=||\mathbf{WX}-\mathbf{(W \odot M) X}||_2^2.
\label{eq:layererror2}
\end{equation}
The goal is to find an optimal sparsity mask $\mathbf{M}$ that reduces model complexity while preserving predictive accuracy.

\textbf{Taylor Expansion at Dense Networks.} For a dense network $\mathbf{\widehat{W}_\text{dense}}$ at a local minimum, the reconstruction error can be expanded into its Taylor series with respect to $\mathbf{\widehat{W}}$, ignoring terms beyond second order:
\begin{equation}
\mathbf{E} = \mathbf{E_d} + \frac{\partial \mathbf{E}}{\partial \mathbf{W}} \Delta \mathbf{W}  + \frac{1}{2} \Delta \mathbf{W}^T \mathbf{H} \Delta \mathbf{W}.
\label{eq:TaylorAtDense}
\end{equation}
Without a sparsity mask in dense model, we can simply assign all-one matrix to the mask M, thereby yielding the zeroth-order terms $\mathbf{E_d} = 0$. The first-order derivative ${\partial \mathbf{E}}/\partial \mathbf{W}$ vanishes when training converged \cite{frantar2023sparsegpt, hassibi1993optimal}, leaving only the computationally expensive second-order terms involving a large Hessian matrix which are challenging for layer-wise reconstruction and channel-wise independence assumption.

\subsection{Block-Aware Rebuilder for Sparsity Mask}
In this work, we depart from existing post-training pruning methods in three key aspects: 
Firstly, to address the layer-aware reconstruction problem that leads to exponentially accumulating biases, we adopt a block-aware reconstruction error and apply a divide-and-conquer strategy to mitigate errors and computational costs.
Secondly, inspired by the pruning-and-growing\footnote{``Pruning'' and ``Growing'' will both be used in following text to refer to rebuild sparsity masks. Pruning changes masks from 1 to 0, whereas growing changes them from 0 to 1.} operation \cite{mocanu2018scalable,zhang2024dynamic}, we address the limitations of mask selection of pruning in dense networks due to the changeable significance of weights, by re-evaluating weight importance score in sparse networks and rebuilding the sparsity mask through targeted growth of salient weights and pruning of non-salient weights. 
Thirdly, our analysis reveals that with the advancement of LLMs, mask selection becomes increasingly critical in weight reconstruction. Thus, we prioritize the rebuilding of the sparsity mask and strip the reconstruction of weights.
Base on these insights, we propose \textbf{LLM-Barber}, a \textbf{Block-Aware Rebuilder for Sparsity Mask in One-Shot} without any need for fine-tuning or retraining.

\textbf{Block-Aware Reconstruction Error.} Building on the definitions in Eq.~(\ref{eq:layererror1}) and Eq.~(\ref{eq:layererror2}), we define the block-aware reconstruction error for a Self-Attention or MLP block:
\begin{equation}
\mathbf{E} = || \mathbf{\text{Block}(W,X)} - \mathbf{\text{Block}(W \odot M, X)} ||_2^2.
\end{equation}
Evaluating reconstruction error across blocks, denoted as $\mathbf{\text{Block}(\cdot)}$, allows us to achieve a globally optimal solution in Self-Attention and MLP blocks rather than layer-wise. This new block-aware reconstruction formulation offers a more effective and cohesive strategy for selecting sparsity masks.

\textbf{Taylor Expansion at Sparse Networks.} Migrating Eq.~(\ref{eq:TaylorAtDense}) at sparse networks $\mathbf{\widehat{W}_{\text{sparse}}}$ with an initialization sparsity mask $\mathbf{M_{i}}$, we can obtain the Taylor series expansion as:
\begin{equation}
\mathbf{E}= \mathbf{E_s} + \mathbf{\frac{\partial E}{\partial W}} \Delta \mathbf{W} + \frac{1}{2} \Delta \mathbf{W}^T \mathbf{H} \Delta \mathbf{W}. 
\end{equation}

The zeroth-order term $\mathbf{E_s}$ of the Taylor expansion at sparse networks represents the reconstruction error after mask initialization:
\begin{equation}
\mathbf{E_s} = || \mathbf{\text{Block}(W, X)} -\mathbf{\text{Block}(W \odot M_i, X)} ||_2^2.
\end{equation}

Assuming non-negligible zeroth-order terms, the first-order gradient in sparse networks remains significant even after convergence and can be efficiently accessed via PyTorch’s Autograd. First-order information provides computational efficiency and operates independently of any reconstruction error. Therefore, second-order terms can be omitted when significant gradients are present, leading to the following change in reconstruction error due to mask rebuilding: 
\begin{equation}
\Delta \mathbf{E}= (\partial \mathbf{E}/\partial \mathbf{W}) \cdot \Delta \mathbf{W},
\end{equation}
which delineates the importance score of weights during sparsity mask rebuilding.

\textbf{Pruning Metric.} 
For a Self-Attention or MLP Block with weights $\mathbf{W}$, first-order information suffices for block-aware reconstruction error in sparse networks. The change in weight magnitude during sparsity mask adjustment matches the weight’s original magnitude ($|\Delta \mathbf{W}{ij}| = |\mathbf{W}{ij}|$). We thus assess the impact of mask rebuilding on reconstruction error by computing the product of the weight’s magnitude and its gradient. The \textbf{importance score} for $\mathbf{W}_{ij}$ is:
\begin{equation}
\mathbf{S}_{ij} = |\mathbf{W}_{ij}| \cdot | (\partial \mathbf{E} / \partial \mathbf{W}_{ij}) |,
\end{equation}
where $|\cdot|$ represents the absolute value operator, and $\mathbf{E}$ denotes the block-aware reconstruction error:
\begin{equation}
\mathbf{E} = || \mathbf{\text{Block}(W, X)}- \mathbf{\text{Block}(W \odot M, X)} ||_2^2.
\end{equation}
%
This method prioritizes weights with both substantial magnitudes and significant gradients, allowing for the preservation of critical weights while pruning less important ones.

\textbf{Pruning Granularity.} 
Choosing the right pruning granularity is crucial \cite{sun2023simple}. Traditional methods operate on a layer-wise \cite{han2015learning}, input-wise \cite{frantar2023sparsegpt}, or output-wise \cite{sun2023simple} basis, which mainly address the layer-aware reconstruction problem, known for its output channel independence. Wanda’s output-wise ranking yields superior results compared to other methods. However, in LLM-Barber’s block-aware framework, output channel independence is no longer applicable. Thus, LLM-Barber extends consideration to block-wise granularity, prioritizing all linear layers within a block. Our analysis of the four distinct granularity levels shows that optimal granularity depends on the specific sparse mask initialization, with detailed results discussed in the Ablation Study \ref{subsec:ablation}.

\textbf{Mask Rebuilding.} 
With the block-aware reconstruction error, gradient information, sparsity matrix, and granularity established, we proceed to rebuild the sparsity mask for each layer. Consider a cluster of weights $\mathbf{W^c}$ under a specific sparsity granularity and its corresponding sparsity mask $\mathbf{M^c}$ and pruning metric $\mathbf{S^c}$. We define the \textbf{growing criterion} and the \textbf{pruning criterion}:
\begin{align}
gi,gj = \text{argmax} \ \mathbf{S^c}, \ \text{if} \  \mathbf{M^c}_{gi,gj} = 0, 
\label{eq:growing}\\
pi,pj = \text{argmin} \ \mathbf{S^c}, \ \text{if} \  \mathbf{M^c}_{pi,pj} = 1.
\label{eq:pruning}  
\end{align}
The growing and pruning weights form a \textbf{mask rebuilding pair}, representing the interchange within the sparsity mask. The value of each pair is defined as the difference between the importance scores of the growing and pruning weights.

Our experiments show that LLM-Barber identifies varying proportions of salient weights based on the sparse mask's initialization method. To regulate mask rebuilding, we introduce a hyperparameter $\alpha$, known as the mask rebuilding ratio. The number of mask rebuilding pairs $N$ is calculated as:
\begin{equation}
N = \{ \ i \  | \ \mathbf{S}^\text{grow}_i - \mathbf{S}^\text{prune}_i > 0 \} \cdot \alpha,
\label{eq:Mask Rebuild Num}
\end{equation}
where {$\mathbf{S}^\text{grow} - \mathbf{S}^\text{prune}$} represents the value of the mask rebuilding pairs,  where $\mathbf{S}^\text{grow}$ is arranged in descending order, and $\mathbf{S}^\text{prune}$ in ascending order. The subscript $i$ denotes the number of values that exceeds zero, indicating that the growing weight is more important than the pruning weight.

\subsection{Procedure}
Like most post-training pruning methods \cite{han2015learning, sun2023simple}, LLM-Barber is executed within a single global LLM forward pass, with local backward passes for gradient computation in each block. Figure \ref{fig:Pipeline} illustrates the LLM-Barber workflow, which consists of four stages:

    \textbf{(1) Sparsity Mask Initialization.} Initialize a preliminary sparsity mask from the dense network by a post-training pruning technique. 
    \textbf{(2) Block-aware Reconstruction Error Computation.} We use a block-aware reconstruction error to evaluate the discrepancy between the dense and sparse model outputs.
    \textbf{(3) Back-propagation for Gradients.} Gradients are automatically derived via back-propagation, and the product of weights and gradients serves as the pruning metric. 
    \textbf{(4) Sparsity Mask Rebuilding.} Masks are sorted by pruning metric, unpruned weights in ascending order while pruned weights in descending order. 
    We rebuild the weight masks by growing newly significant weights and pruning those became non-salient with a specific mask rebuilding ratio.

Ultimately, we recalculated the block-aware reconstruction error to assess the enhancement brought by our LLM-Barber. Experimental results reveal a substantial decrease in error, markedly decelerating the rate of error accumulation.

\begin{algorithm}[t]
\textbf{Input}: Calibration samples $\mathbf{X}$, a block's weights $\{\mathbf{W}^l\}^L_{l=1}$ and initial masks $\{\mathbf{M_{i}}^l\}^L_{l=1}$,  mask rebuilding ratio $\alpha$. \\
\textbf{Output}: Rebuilt sparsity masks $\{\mathbf{M_{r}}^l\}^L_{l=1}$. 

\begin{algorithmic}[1]
\STATE $\mathbf{E_i} \gets \text{Block}\mathbf{(W, X)} - \text{Block}\mathbf{(W \odot M_i, X)}$
\STATE $\{\mathbf{G}^l\}^L_{l=1} \gets$ BP for gradients via $\mathbf{E_i}$.
\FOR{$l$ in $\{1,2, \dots, L\}$}
    \STATE $\mathbf{M_r}^l \gets \mathbf{M_i}^l$
    \STATE $\mathbf{S}^l \gets |\mathbf{W}^l| \cdot |\mathbf{G}^l|$
    \STATE $N \gets \mathbf{S}^l, \alpha \  \text{via Eq.} (\ref{eq:Mask Rebuild Num})$
    \FOR{$n$ in $\{1,2, \dots, N\}$}
        \STATE Obtain growing index $gi,gj$ via Eq.(\ref{eq:growing})
        \STATE Obtain pruning index $pi,pj$ via Eq.(\ref{eq:pruning})
        \STATE $\mathbf{M_r}^l_{pi,pj} = 1$ \COMMENT{Weight Growing.}
        \STATE $\mathbf{M_r}^l_{gi,gj} = 0$ \COMMENT{Weight Pruning.}
    \ENDFOR
\ENDFOR
\STATE $\mathbf{E_r} \gets \text{Block}\mathbf{(W, X)} - \text{Block}\mathbf{(W \odot M_r, X)}$
\STATE Identify improvement by comparing $\mathbf{E_i}$, $\mathbf{E_r}$.
\STATE \textbf{return} rebuilt sparsity masks $\{\mathbf{M_{r}}^l\}^L_{l=1}$.
\end{algorithmic}
\caption{Pseudocode of LLM-Barber.}
\end{algorithm}

\subsection{Structured N:M Sparsity}
\label{subsec:nm}
While LLM-Barber primarily targets unstructured sparsity, it can be adapted for structured N:M sparsity. Groups of M weights are pruned to retain only N non-zero weights. 
During mask rebuilding, it divides each M-group into N pairs (one pruned and one non-pruned weight), then sorts them by output channel to identify mask rebuilding pairs. This method optimizes sparsity mask, leveraging N:M sparsity while maintaining model performance.

\begin{table*}[t]
\caption{WikiText-2 perplexity comparison at 50\% sparsity rate. \textbf{Bold} results show improvements of integrating LLM-Barber. \underline{Underscored} results indicate best performance in each LLM.WR represents weight reconstruction.}
    \centering
    \begin{tabular}{c|cc|cc|c|cc}
    \toprule
    & \multicolumn{2}{c|}{LLaMA1}  & \multicolumn{2}{c|}{LLaMA2} & \multicolumn{1}{c|}{ LLaMA3 } & \multicolumn{2}{c}{OPT} \\
    \midrule
    \textbf{Method} 
    & \textbf{7B} & \textbf{13B} & \textbf{7B} & \textbf{13B} & \textbf{8B} & \textbf{6.7B} & \textbf{13B} \\
    \midrule 
    Dense & 5.677 & 5.091 & 5.472 & 4.884 & 6.136 & 10.86 & 10.13 \\
    \midrule
    Magnitude & 17.26 & 20.14 & 16.03 & 6.827 & 205.5 & 9.7e2 & 1.2e4 \\
    \rowcolor{gray!10} \textit{w/} \textbf{LLM-Barber} 
              & \textbf{7.332} & \underline{\textbf{6.089}} & \textbf{7.170} & \textbf{5.955} & \textbf{10.98} & \textbf{13.12} & \textbf{15.52} \\
    \midrule
    SparseGPT & 7.201 & 6.194 & 7.005 & 6.036 & 9.399 & \underline{11.59} & \underline{11.15} \\
    SparseGPT \textit{w/o} WR & 7.545 & 6.311 & 7.413 & 6.134 & 9.994 & 13.13 & 15.76 \\
    \rowcolor{gray!10} \textit{w/} \textbf{LLM-Barber} 
              & \textbf{7.159} & \textbf{6.125} & \textbf{7.004} & \textbf{5.929} & \underline{\textbf{9.348}} & \textbf{11.95} & \textbf{11.93} \\
    \midrule
    Wanda & 7.254 & 6.152 & 6.920 & 5.972 & 9.821 & 11.98 & 11.93 \\
    \rowcolor{gray!10} \textit{w/} \textbf{LLM-Barber} 
              & \underline{\textbf{7.118}} & \textbf{6.091} & \underline{\textbf{6.868}} & \underline{\textbf{5.918}} & \textbf{9.451} & \textbf{11.95} & \textbf{11.71} \\
    \bottomrule
    \end{tabular}
    \captionsetup{justification=raggedright}
    \label{tab:Unstructured Sparsity}
\end{table*}

\section{Experiment}
\label{sec:expt}
\subsection{Experiment Settings}
\textbf{Setup.} LLM-Barber is implemented in Pytorch and utilized public model checkpoints from the HuggingFace library \footnote{huggingface.co/meta-llama, huggingface.co/huggyllama} on a single 80GB NVIDIA A100 GPU. After mask initilization, LLM-Barber uniformly rebuilds sparsity masks in sequence, performing in one-shot without any fine-tuning.

\textbf{Models \& Datasets.} 
LLM-Barber is evaluated on the LLaMA family, including LLaMA-7B/13B \cite{touvron2023llama}, LLaMA2-7B/13B \cite{touvron2023llama2}, and LLaMA3-8B \cite{llama3modelcard}, as well as the OPT model series: OPT-6.7B/13B \cite{zhang2022opt}. Notably, LLM-Barber is broadly applicable to any Transformer-based LLMs with Self-Attention and MLP blocks. Following previous works, we use 128 segments of 2048 tokens from the C4 dataset \cite{raffel2020exploring} for mask rebuilding.

\textbf{Evaluation.} To comprehensively assess LLM-Barber, we conduct rigorous evaluations on \textit{perplexity} and \textit{zero-shot accuracy}.
Perplexity is measured on the validation sets of benchmarks such as WikiText-2 \cite{Merity2016wikitext}, PTB \cite{Marcus1994ptb}, and C4 \cite{raffel2020exploring}. 
Zero-shot accuracy is assessed using the EleutherAI LM Harness~\cite{eval-harness} across six benchmarks: BoolQ \cite{BoolQ}, RTE \cite{RTE}, HellaSwag \cite{HellaSwag}, ARC Easy and Challenge \cite{ARC}, and OpenbookQA \cite{OpenBookQA}.

\textbf{Baselines.} 
The results of LLM-Barber are compared with the following established post-training pruning methods: (1) Magnitude Pruning \cite{han2015learning} eliminates weights based only on their magnitudes; (2) SparseGPT \cite{frantar2023sparsegpt} identifies weights importance by using second-order information; (3) Wanda \cite{sun2023simple} determines weights to be pruned by the weight magnitude multiplied by input activation.

\subsection{Main Results}
\textbf{Unstructured Sparsity.} LLM-Barber effectively prunes the LLaMA and OPT models, achieving 50\% unstructured sparsity without requiring supplementary weight reconstruction, as detailed in Table~\ref{tab:Unstructured Sparsity}. LLM-Barber demonstrates the capability to rebuild the sparsity masks initialized by other pruning methods in a single forward pass, significantly outperforming conventional pruning baselines.
In the LLaMA3-8B model, LLM-Barber creates a new sparsity mask that reduces perplexity to 9.451, a substantial improvement over the Wanda baseline of 9.821.
Notably, LLM-Barber achieves robust improvements even with poorly performing initial sparsity masks, such as magnitude pruning, where it reduces perplexity from 205.5 to 10.98, which is an impressive and substantial enhancement.

\textbf{Varying Sparsity Levels.} We conduct experiments on varying sparsity levels for unstructured pruning in LLaMA3-8B as shown in Table \ref{tab:Varying Sparsity Levels}. LLM-Barber consistently increases perplexity across all initialization methods, with magnitude pruning showing the most significant improvement.

\begin{table}[t]
    \centering
    \captionsetup{justification=raggedright}
    \caption{WikiText-2 perplexity performance for pruning LLaMA3-8B at varying sparsity rate.} 
    \label{tab:Varying Sparsity Levels}
    \begin{tabular}{c|cccc}
    \toprule
    \textbf{Sparsity} 
    & \textbf{60\%} & \textbf{70\%} & \textbf{80\%} & \textbf{90\%} \\
    \midrule 
    Magnitude & 3.39e4 & 1.62e6 & 8.55e7 & 2.26e7 \\
    \rowcolor{gray!10} \textit{w/} \textbf{Barber} 
            & {\textbf{28.14}} & \textbf{2.08e2} & \textbf{1.09e3} & \textbf{7.55e4} \\
    \midrule
    Wanda &  23.57 & 1.28e2 & 8.54e2 & 1.28e4 \\
    \rowcolor{gray!10} \textit{w/} \textbf{Barber} 
               & \textbf{22.04} & {\textbf{1.07e2}} & {\textbf{5.70e2}} & \textbf{7.07e3} \\
    \bottomrule
    \end{tabular}
\end{table}

\textbf{Structured N:M Sparsity.} 
In contrast to unstructured sparsity, employing N:M fine-grained sparsity can provide more tangible acceleration benefits when leveraging NVIDIA Ampere's sparse tensor cores~\cite{choquette2021nvidia}. Therefore, we also evaluate the effectiveness of our LLM-Barber in partial LLaMA models on the N:M fine-grained sparsity pattern as shown in Table~\ref{tab: Structured N:M Sparsity.}. 

\begin{table}[t]
    \centering
    \captionsetup{justification=raggedright}
    \caption{WikiText-2 perplexity comparison for pruning LLaMA family with structured N:M pattern.}
    \begin{tabular}{c|c|ccc}
    \toprule
    \textbf{Method} 
    & \textbf{Sparsity} & \textbf{V1-7B} & \textbf{V1-13B} & \textbf{V2-7B} \\
    \midrule 
    Magnitude & 4:8 & 16.83 & 13.72 & 15.91 \\
    \rowcolor{gray!10} \textit{w/} \textbf{Barber} & 4:8
              & \textbf{8.852} & {\textbf{7.137}} & \textbf{9.345} \\
    \midrule
    SparseGPT & 4:8 & 8.608 & 7.437 & 8.495 \\
    \rowcolor{gray!10} \textit{w/} \textbf{Barber} & 4:8
              & {\textbf{8.191}} & \textbf{7.085} & {\textbf{8.003}} \\
    \midrule 
    Magnitude & 2:4 & 42.56 & 18.32 & 37.76 \\
    \rowcolor{gray!10} \textit{w/} \textbf{Barber} & 2:4
              & \textbf{11.04} & {\textbf{9.006}} & \textbf{13.47} \\
    \midrule
    SparseGPT & 2:4 & 11.55 & 9.116 & 10.94 \\
    \rowcolor{gray!10} \textit{w/} \textbf{Barber} & 2:4
              & {\textbf{10.14}} & \textbf{8.517} & {\textbf{9.806}} \\
    \bottomrule
    \end{tabular}
    \label{tab: Structured N:M Sparsity.}
\end{table}

\begin{table*}[ht]
    \centering
    \captionsetup{justification=raggedright}
    \caption{Zero-shot performance comparison of LLaMA series on six tasks at 50\% sparsity rate. \textbf{Bold} results show improvements of integrating LLM-Barber. \underline{Underscored} results indicate the best performance for each tasks.}
    \begin{tabular}{c|c|ccccccc}
    \toprule
    \textbf{Model} & \textbf{Method} 
    & \textbf{BoolQ} & \textbf{RTE} & \textbf{HellaSwag} & \textbf{ARC-e} & \textbf{ARC-c} & \textbf{OBQA} & \textbf{Mean}\\
    \midrule 
    \multirow{5}{*}{LLaMA-7B} & Dense & 75.08 & 66.79 & 56.96 & 75.29 & 41.89 & 34.40 & 58.40 \\
    \cmidrule{2-9}
    & Magnitude & 54.61 & 54.51 & 45.47 & 58.75 & 33.45 & 22.60 & 44.89 \\
    &\cellcolor{gray!10}\textit{w/} \textbf{LLM-Barber} &\cellcolor{gray!10}\textbf{71.47} &\cellcolor{gray!10}\textbf{\underline{59.93}} &\cellcolor{gray!10}\textbf{50.63} &\cellcolor{gray!10}\textbf{69.11} &\cellcolor{gray!10}\textbf{35.41} &\cellcolor{gray!10}\textbf{28.20} &\cellcolor{gray!10}\textbf{52.46} \\
    & Wanda & 70.98 & 55.23 & \underline{51.90} & 69.44 & 36.86  & 28.60 & 52.17 \\
    &\cellcolor{gray!10}\textit{w/}\textbf{LLM-Barber} &\cellcolor{gray!10}\underline{\textbf{73.12}} &\cellcolor{gray!10}\textbf{56.68} &\cellcolor{gray!10}51.80 &\cellcolor{gray!10}\underline{\textbf{70.32}} &\cellcolor{gray!10}\underline{\textbf{36.95}} &\cellcolor{gray!10}\underline{\textbf{28.80}} &\cellcolor{gray!10}\underline{\textbf{52.95}} \\
    \midrule 
    \multirow{5}{*}{LLaMA2-13B} & Dense & 80.58 & 65.34 & 60.05 & 79.38 & 48.46 & 35.20 & 61.50 \\
    \cmidrule{2-9}
    & Magnitude & 70.53 & 55.96 & 54.42 & 57.68 & 38.40 & 27.80 & 50.79\\
    &\cellcolor{gray!10}\textit{w/} \textbf{LLM-Barber} 
              &\cellcolor{gray!10}\underline{\textbf{81.10}} &\cellcolor{gray!10}\textbf{60.29} &\cellcolor{gray!10}\textbf{55.67} &\cellcolor{gray!10}\textbf{75.34} &\cellcolor{gray!10}\textbf{40.28} &\cellcolor{gray!10}\textbf{31.20} &\cellcolor{gray!10}\textbf{57.31} \\
    & Wanda & 80.95 & 60.28 & \underline{56.98}     & 76.30 & \underline{42.26} & 31.20 & 57.99 \\
    &\cellcolor{gray!10}\textit{w/} \textbf{LLM-Barber} 
              &\cellcolor{gray!10}\textbf{80.98} &\cellcolor{gray!10}\underline{\textbf{62.82}} &\cellcolor{gray!10}55.99 &\cellcolor{gray!10}\underline{\textbf{76.39}} &\cellcolor{gray!10}42.13 &\cellcolor{gray!10}\underline{\textbf{31.40}} &\cellcolor{gray!10}\underline{\textbf{58.29}} \\
    \midrule 
    \multirow{5}{*}{LLaMA3-8B} & Dense & 81.35 & 69.68 & 60.19 & 80.09 & 50.60 & 34.80 & 62.79  \\
    \cmidrule{2-9}
    & Magnitude & 42.87 & 53.07 & 29.85 & 46.59 & 25.09 & 22.00 & 36.57\\
    &\cellcolor{gray!10}\textit{w/} \textbf{LLM-Barber} 
              &\cellcolor{gray!10}\textbf{72.72} &\cellcolor{gray!10}\textbf{54.87} &\cellcolor{gray!10}\textbf{51.00} &\cellcolor{gray!10}\underline{\textbf{72.05}} &\cellcolor{gray!10}\textbf{37.46} &\cellcolor{gray!10}\textbf{27.40} &\cellcolor{gray!10}\textbf{52.58}\\
    & Wanda & 78.41 & 60.29 & \underline{51.20} & 71.38 & \underline{40.10} & \underline{29.40} & 55.13\\
    & \cellcolor{gray!10}\textit{w/}\textbf{LLM-Barber} 
              &\cellcolor{gray!10}\underline{\textbf{78.59}} &\cellcolor{gray!10}\underline{\textbf{61.37}} &\cellcolor{gray!10}51.01 &\cellcolor{gray!10}\textbf{71.46} &\cellcolor{gray!10}39.85 &\cellcolor{gray!10}29.00 &\cellcolor{gray!10}\underline{\textbf{55.21}} \\
    \bottomrule
    \end{tabular}
    \label{tab:Zeroshot Performance}
\end{table*}
\textbf{Zero-shot Performance.} Following previous works \cite{frantar2023sparsegpt, sun2023simple}, we evaluated the LLaMA models on six diverse zero-shot tasks. The results are summarized in Table~\ref{tab:Zeroshot Performance}, where models are pruned to unstructured 50\% sparsity.
Averaging the accuracy across the six evaluated tasks, it becomes apparent that LLM-Barber possesses the capability to identify a more effective network than those obtained via the initialization methods. 
For tasks such as BoolQ, RTE, and ARC-e, LLM-Barber consistently outperforms the baseline pruning techniques across the entire LLaMA model suite. However, it is worth noting that there is no single universally superior performer for the remaining tasks in the evaluation set, with the initial pruning methods sometimes matching or even marginally exceeding the results obtained through LLM-Barber.

\begin{table*}[t]
\centering
\caption{Comparison of DSnoT vs. LLM-Barber on LLaMA3-8B at 50\% unstructured sparsity. We report WikiText-2 perplexity (lower is better) and zero-shot accuracy (\%) on five downstream tasks. Bold indicates best result for each setting.}
\label{tab:dsnot}
\begin{tabular}{lcccccc}
\toprule
\textbf{Method} & \textbf{WikiText-2} & \textbf{ARC-E} & \textbf{ARC-C} & \textbf{BoolQ} & \textbf{OBQA} & \textbf{RTE} \\
\midrule
Dense (no pruning) & 6.136 & 75.29 & 41.89 & 75.08 & 34.40 & 66.79 \\
Magnitude (50\%) & 205.5 & 46.59 & 25.09 & 42.87 & 22.00 & 53.07 \\
\ \ w/ DSnoT & 47.52 & 60.52 & 32.16 & 49.66 & 25.20 & 54.87 \\
\ \ \textbf{w/ LLM-Barber} & \textbf{10.98} & \textbf{72.05} & \textbf{37.49} & \textbf{72.72} & \textbf{27.40} & \textbf{54.87} \\
\midrule
Wanda (50\%) & 9.821 & 71.38 & 40.10 & 78.41 & 29.40 & 60.29 \\
\ \ w/ DSnoT & 9.562 & 71.25 & 40.01 & 75.99 & 29.00 & \textbf{61.81} \\
\ \ \textbf{w/ LLM-Barber} & \textbf{9.451} & \textbf{71.46} & \textbf{39.85} & \textbf{78.59} & \textbf{29.00} & 61.37 \\
\bottomrule
\end{tabular}
\end{table*}

\textbf{Comparison with DSnoT.} 
We further compare LLM-Barber with the recent training-free pruning method DSnoT~\cite{zhang2024dynamic} to highlight their differences in approach and performance. DSnoT also aims to improve sparse LLMs without weight retraining, but it does so by \emph{iteratively} pruning and regrowing weights in multiple cycles to minimize reconstruction In contrast, LLM-Barber performs all mask rebuilding in a single global forward pass (with localized backward gradient computations per block), making it a true one-shot method. This fundamental distinction gives LLM-Barber an efficiency edge: it avoids the repeated passes of DSnoT, enabling faster one-shot deployment. Moreover, DSnoT imposes a fixed number of iterations (a maximum cycle count) for mask update in each layer, which can lead to sub-optimal results when the initial mask is poor. LLM-Barber is not bound by a preset iteration limit; instead, it dynamically rebuilds the mask based on our weight-importance metric until all significant weights are retained. This adaptability allows LLM-Barber to recover much more performance, especially starting from an inferior initial mask. Importantly, LLM-Barber’s \emph{block-aware} strategy provides greater flexibility via \textit{global coordination} across Transformer blocks, whereas DSnoT treats layers more independently (resembling layerwise fine-tuning). By coordinating mask updates at the block level, LLM-Barber better preserves end-to-end model quality. 

In summary, qualitatively LLM-Barber offers higher efficiency, flexibility, and overall performance than DSnoT. Its one-shot block-aware mask rebuilding avoids DSnoT’s iterative overhead and limitation, yielding a more globally coordinated pruning outcome. These advantages make LLM-Barber well-suited for scenarios that require rapid one-shot compression for LLM deployment without sacrificing accuracy.

\textbf{Broader Downstream Tasks.} 
While our evaluation in Table~\ref{tab:Zeroshot Performance} and Table~\ref{tab:dsnot} spans a diverse set of zero-shot tasks, we recognize the need to demonstrate the effectiveness of LLM-Barber on an even broader range of downstream applications. In particular, many tasks in the GLUE benchmark (e.g., SST-2 sentiment analysis, MNLI textual entailment) and other NLP benchmarks remain to be fully tested under our one-shot pruning framework. We conducted preliminary experiments on several GLUE tasks and observed that pruned models maintain competitive accuracy relative to dense models. For example, on the SST-2 dataset, a LLaMA2-7B model pruned to 50\% with LLM-Barber still achieves a $92.1\%$ accuracy, only a small drop from the $93.6\%$ achieved by the dense model (zero-shot classification setting). This suggests that LLM-Barber’s benefits extend to classification tasks with minimal performance loss. A comprehensive evaluation on the entire GLUE suite is underway – due to computational constraints, not all results were ready at the time of writing – and will be included in a future extension of this work. These ongoing efforts will further verify that LLM-Barber generalizes well across many downstream tasks, addressing the concern about its applicability beyond the benchmarks reported here.

\subsection{Mask Rebuilding Ratio Selection}
\begin{figure*}[ht]
    \centering
    \begin{minipage}{0.32\textwidth}
        \includegraphics[width=\textwidth]{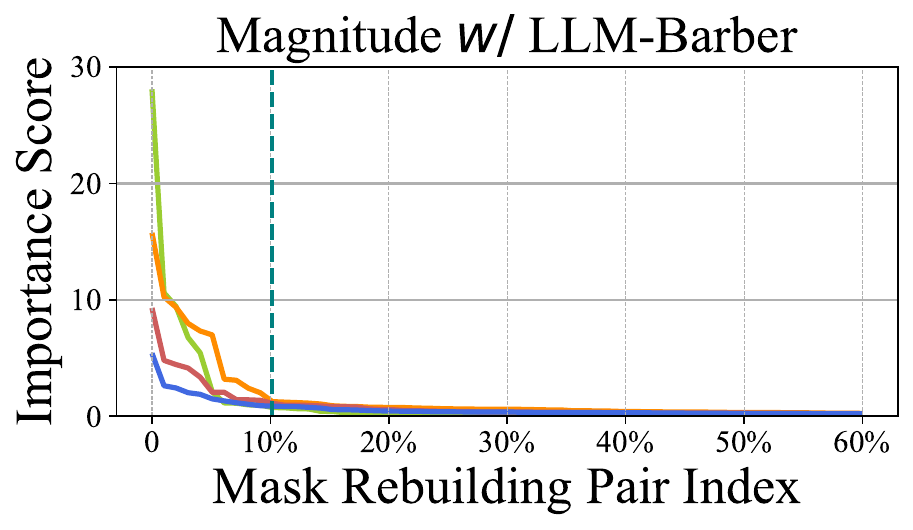}
    \end{minipage}
    \begin{minipage}{0.32\textwidth}
        \includegraphics[width=\textwidth]{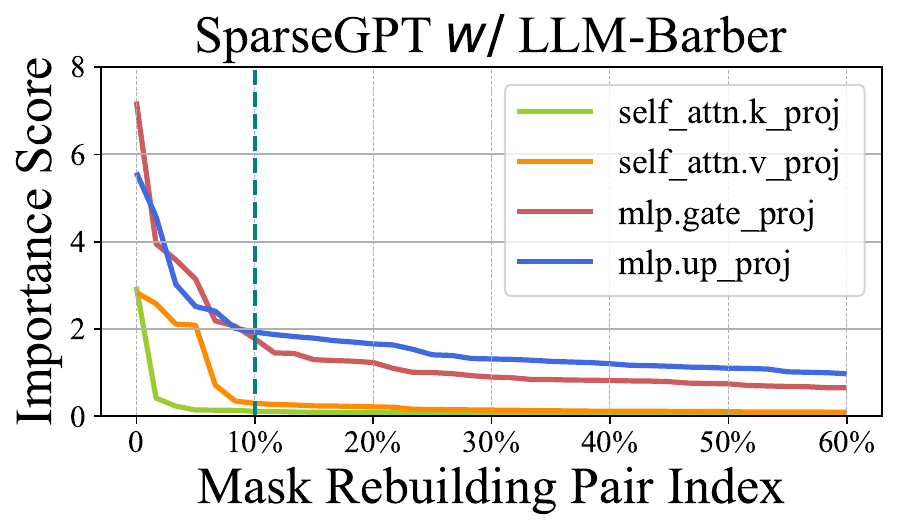}
    \end{minipage}
    \begin{minipage}{0.32\textwidth}
        \includegraphics[width=\textwidth]{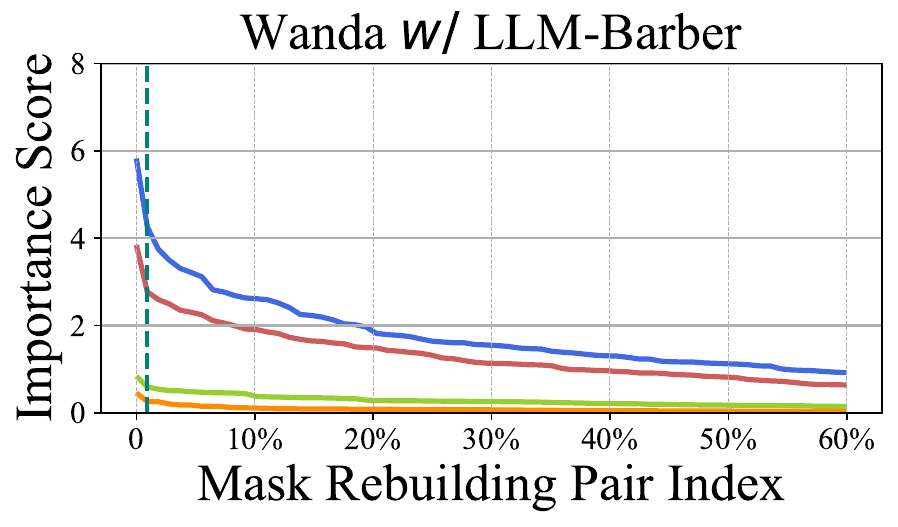}
    \end{minipage}
    
    \centering
    \begin{minipage}{0.32\textwidth}
        \includegraphics[width=\textwidth]{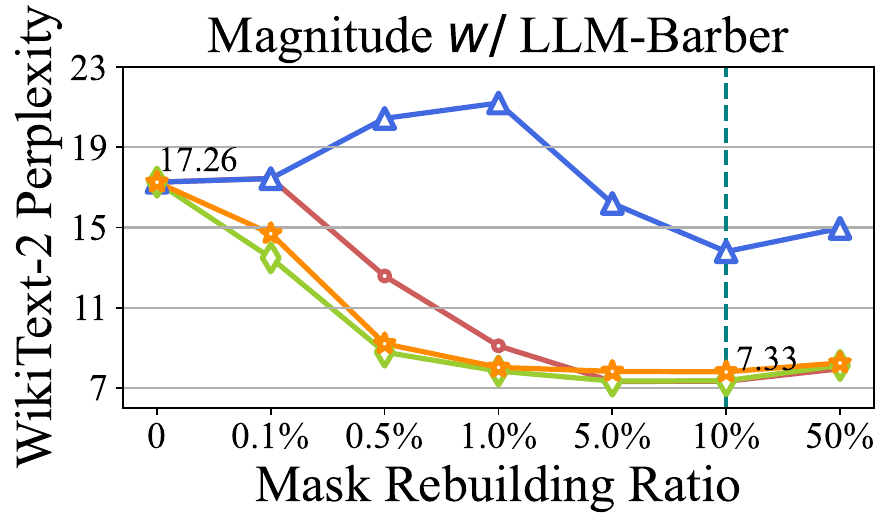}
    \end{minipage}
    \begin{minipage}{0.32\textwidth}
        \includegraphics[width=\textwidth]{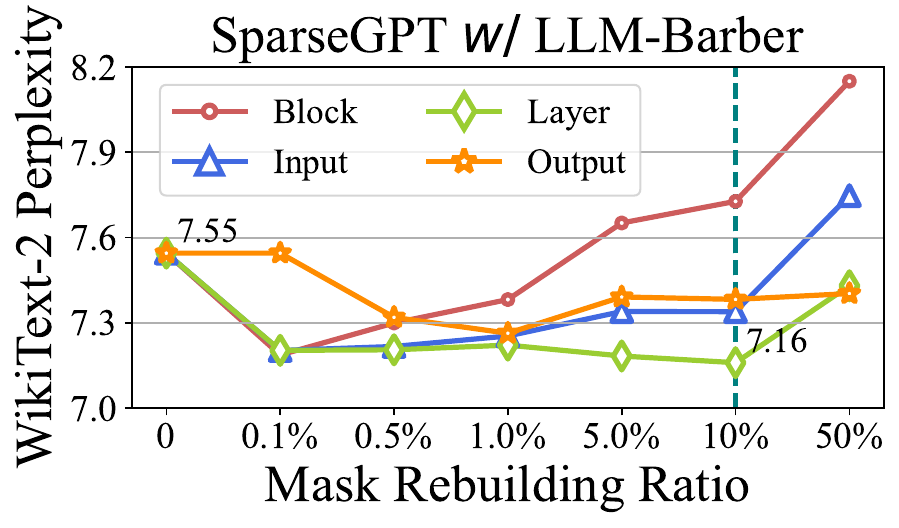}
    \end{minipage}
    \begin{minipage}{0.32\textwidth}
        \includegraphics[width=\textwidth]{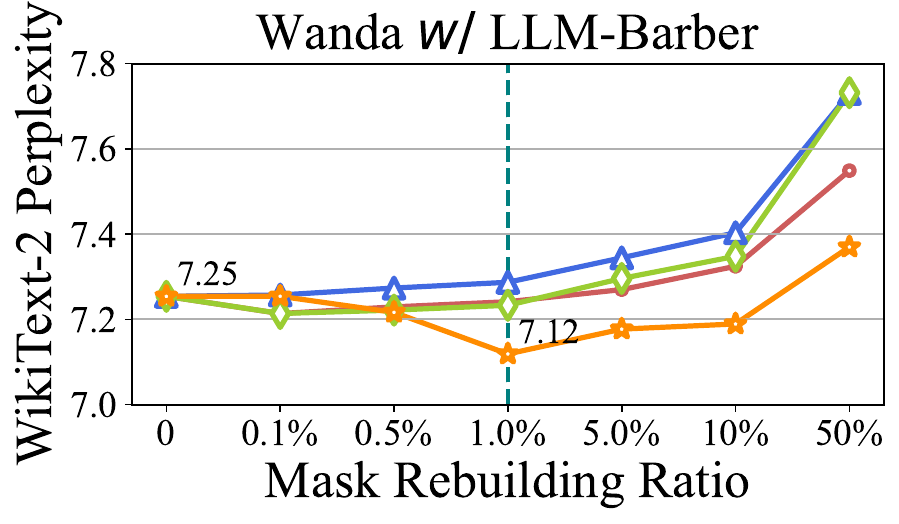}
    \end{minipage}
    \captionsetup{justification=raggedright}
    \caption{The importance score distribution of mask rebuilding pairs and WikiText-2 perplexity results at varying pruning granularities in LLaMA-7B, with the green dashed line marking the optimal mask rebuilding ratio.}
    \label{fig:pdf_files}
\end{figure*}
A critical aspect of the LLM-Barber method lies in determining the optimal mask rebuilding ratio to achieve peak accuracy. One effective strategy involves analyzing the distribution of value magnitudes within the mask rebuilding pairs, corresponding to differences between the growing and pruning importance score. Fortunately, this distribution often exhibits a distinct pattern of outliers, facilitating rapid identification of an appropriate mask rebuilding ratio.

We plotted the score distribution of mask rebuilding pairs across various pruning granularity and initialization methods on LLaMA-7B, along with perplexity performance corresponding to mask rebuilding ratios (Figure~\ref{fig:pdf_files}). The results reveal a strong correlation between the distribution of outliers in mask rebuilding pairs and the optimal mask rebuilding ratio. For instance, outliers are significantly distributed within the top 10\% with Magnitude pruning, thus selecting a 10\% mask rebuilding ratio yields the overall optimal solution. 
Similarly, the Wanda initialization method shows a notable outlier distribution around the 1\% mark corresponds to optimal results near a 1\% mask rebuilding ratio. Thus, LLM-Barber can preemptively narrow the search range for the optimal mask rebuilding ratio by analyzing outlier distributions, allowing for flexible adaptation to various reconstruction masks without extensive searching.

It is worth noting that LLM-Barber identifies a larger proportion of outliers in less effective initial masks, which corresponds to a more substantial rebuilding of the mask. This is why LLM-Barber provides a more significant improvement for masks with poorer initial sparsity mask with a more aggressive mask rebuilding strategy.

\begin{table}[htbp]
    \centering
    \captionsetup{justification=raggedright}
    \caption{Pruning granularity ablation in LLaMA3-8B. \textbf{Bold} results show best granularity of each row.}
    \begin{tabular}{c|cccc}
    \toprule
    \textbf{Method} & \multicolumn{4}{c}{\textbf{Pruning Granularity}} \\ 
    \rowcolor{gray!10} \textit{w/} \textbf{Barber} & \textbf{Block} & \textbf{Layer} & \textbf{Input} & \textbf{Output} \\
    \midrule 
    Magnitude & \textbf{10.98} & 11.06 & 72.48 & 11.81 \\
    SparseGPT & 9.380 & \textbf{9.348} & 9.418 & 9.567 \\
    Wanda & 9.626 & 9.633 & 9.849 & \textbf{9.451 }\\
    \bottomrule
    \end{tabular}
    \label{tab: Pruning Granularity.}
\end{table}
\subsection{Ablation Study}
\label{subsec:ablation}
Given the significant potential of LLM-Barber, we analyzed three critical factors to assess the robustness and effectiveness of our method: pruning granularity, pruning metric, and calibration data size. These factors were chosen to gain a deeper understanding of how different configurations affect the performance of pruned models and to demonstrate LLM-Barber's versatility across various settings.

\begin{table}[htbp]
    \centering
    \captionsetup{justification=raggedright}
    \caption{Ablation of pruning metric in LLaMA3-8B. \textbf{Bold} results support the pruning metric $|\mathbf{W}| |\mathbf{\partial E / \partial W}|$.}
    \begin{tabular}{c|ccc}
    \toprule
    \textbf{Method} & \multicolumn{3}{c}{\textbf{Pruning Metric}} \\ 
    \rowcolor{gray!10} \textit{w/} \textbf{Barber} & { $|\mathbf{W}|$} & {$|\mathbf{\partial E / \partial W}|$} & {$|\mathbf{W}| |\mathbf{\partial E / \partial W}|$} \\
    \midrule 
    Magnitude & 186.5 & 14.43 & \textbf{10.98}  \\
    SparseGPT & 9.544 & 9.457 & \textbf{9.348}  \\
    Wanda & 9.880 & 9.699 & \textbf{9.451} \\
    \bottomrule
    \end{tabular}
    \label{tab: Pruning Metric.}
\end{table}
\textbf{Pruning Granularity.} 
LLM-Barber dynamically selects different pruning granularities to adapt to varying initialization methods. In this paper, we evaluated the impact of four levels of granularities: block-wise, layer-wise, input-wise, and output-wise pruning, as shown in Table \ref{tab: Pruning Granularity.}.
For magnitude pruning, the block-wise granularity yields the best performance, while the output-wise granularity delivers the lowest perplexity for Wanda pruning. By tailoring the pruning granularity to the particular pruning approach, LLM-Barber can consistently achieve optimal model performance.

\begin{figure}[htbp]
  \centering
  \includegraphics[width=0.9\linewidth]{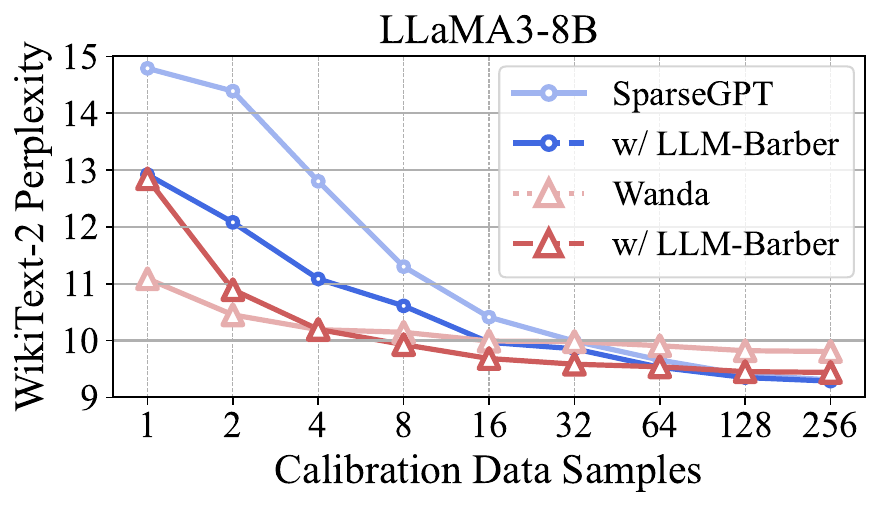}
  \captionsetup{justification=raggedright}
  \caption{Ablation of calibration size in LLaMA3-8B. LLM-Barber is robust across varying calibration size.}
  \label{fig:calibration}
\end{figure}

\begin{table*}[t]
\small\centering
\caption{Inference performance of different quantization methods on FPGA platform. Results are reported for Qwen2-7B on EdgeLLM with the ARC-Challenge, ARC-Easy, and BoolQ tasks in zero-shot setting.}
\begin{tabular}{l|ccc|ccc|ccc}
\hline
\multirow{2}{*}{Method} & \multicolumn{3}{c|}{ARC-Challenge} & \multicolumn{3}{c|}{ARC-Easy} & \multicolumn{3}{c}{BoolQ} \\
 & TPS $\uparrow$ & Latency(ms) $\downarrow$ & Acc(\%) $\uparrow$ & TPS $\uparrow$ & Latency(ms) $\downarrow$ & Acc(\%) $\uparrow$ & TPS $\uparrow$ & Latency(ms) $\downarrow$ & Acc(\%) $\uparrow$ \\
\hline
GPTQ & 30.01 & 1.98 & 46.28 & 29.88 & 1.75 & 64.71 & 30.31 & 0.65 & 56.88 \\
AWQ  & 28.13 & 2.23 & 35.38 & 28.85 & 2.02 & 57.62 & 28.47 & 1.98 & 51.84 \\
APTQ+Ours & \textbf{30.37} & \textbf{1.92}  & \textbf{56.28} & \textbf{30.14} & \textbf{1.58}  & \textbf{67.16} & \textbf{31.48} & \textbf{0.48} & \textbf{60.94} \\
\hline
\end{tabular}
\label{tab:fpga_results}
\end{table*}
\textbf{Pruning Metric.}
We analyze the effect of different pruning metrics, weight magnitude, gradient magnitude, and product of weight and gradient. As shown in Table~\ref{tab: Pruning Metric.}, the product of weight and gradient consistently outperforms the others, achieving the lowest perplexity of 10.98 for Magnitude, 9.348 for SparseGPT, and 9.451 for Wanda. This confirms the effectiveness of our product-based pruning metric across various methods.

\textbf{Calibration Data Size.}
We explore how varying the size of calibration data influences the performance of LLM-Barber. Figure~\ref{fig:calibration} demonstrates that as the calibration sample size increases, LLM-Barber maintains robust performance. Notably, with more than four samples, our method achieves better perplexity than SparseGPT and Wanda. LLM-Barber outperforms SparseGPT even with just a single sample, underscoring its robustness across different sample sizes.

\subsection{Deployment-Aware Quantization Compatibility}
\begin{table}[htbp]
\small
\centering
\caption{Comparative Throughput Analysis of Quantization Methods Across GPU Frameworks with LLM-Barber Optimization}
\begin{tabular}{l|cccc|c}
\hline
Method & \multicolumn{4}{c|}{Throughput(Tokens/sec)} & Baseline \\
 & CUDA & Triton & Marlin & ExLlama & BF16 \\
\hline
GPTQ & 37.38 & 15.90 & 38.03 & 50.59 & 44.30 \\
AWQ  & 38.10 & 16.21 & 43.46 & 51.63 & 44.30 \\
APTQ+Ours & \textbf{38.29} & \textbf{16.42} & \textbf{47.13} & \textbf{52.29} & 44.30 \\
\hline
\end{tabular}
\label{tab:gpu_results}
\end{table}

LLM-Barber's block-aware sparsity method provides an efficient compression solution when combined with post-training quantization like APTQ~\cite{guan2024aptq}. This integration enhances both model size reduction and inference speed. In this section, we present experimental results validating the compatibility and acceleration benefits of combining LLM-Barber with APTQ.

We evaluate the model's performance on both GPU and FPGA platforms, comparing tokens per second (TPS), latency, and accuracy across different quantization methods: GPTQ~\cite{frantar-gptq}, AWQ~\cite{lin2023awq}, and APTQ. As shown in Table~\ref{tab:gpu_results}, LLM-Barber combined with APTQ achieves higher TPS and lower latency on both platforms, while maintaining competitive accuracy.

Additionally, we extend this analysis to FPGA hardware using the EdgeLLM~\cite{huang2025edgellm} platform, where we compare the performance of the quantization methods on the ARC-Challenge, ARC-Easy, and BoolQ datasets. As detailed in Table~\ref{tab:fpga_results}, APTQ, when integrated with LLM-Barber's block-sparsity, delivers the highest TPS, lowest latency, and best accuracy across all tasks. This validates the effectiveness of LLM-Barber’s block-aware sparsity and APTQ’s quantization on FPGA, demonstrating that our method significantly accelerates deployment while preserving model performance.

In summary, the combination of LLM-Barber's block-sparsity approach with APTQ optimizes both inference speed and deployment flexibility. Our results confirm that this joint optimization provides higher TPS and lower latency on GPU while achieving substantial acceleration on FPGA platforms. These findings highlight the potential of integrating sparse pruning with quantization for end-to-end deployment of LLMs, enabling scalable, efficient, and hardware-agnostic solutions across different platforms, such as GPU and FPGA.

Future work will explore further quantization schemes and optimizations for edge deployment. We envision expanding this approach to larger model architectures and more diverse hardware environments, making large-scale LLM deployment more efficient and accessible.

\section{Conclusion}
We introduced LLM-Barber, a one-shot pruning framework that achieves aggressive compression of large language models without retraining. By leveraging a block-aware mask rebuilding strategy and a weight-gradient importance metric, LLM-Barber removes 50–60\% of weights while preserving state-of-the-art accuracy. Our approach prunes multi-billion-parameter models in minutes and incorporates hardware-aware optimizations, such as structured block sparsity and low-bit quantization, to ensure effective deployment on modern accelerators. Experimental results demonstrate that LLM-Barber outperforms unstructured pruning baselines like SparseGPT and Wanda in both perplexity and downstream task accuracy. Additionally, the FPGA deployment of a pruned+quantized model achieves significant speedups and energy savings, proving LLM-Barber's hardware efficiency. In conclusion, LLM-Barber bridges the gap between theoretical model compression and practical deployment, enabling efficient LLM inference in resource-constrained environments. Future work will scale this approach to larger models and integrate it with other compression techniques.

\section*{Acknowledgement}
This work was supported by the Shenzhen Science and Technology Program under Grant No. KQTD20200820113051096, 
Shenzhen Science and Technology Program under Grant No. JCYJ20220818100217038,
the National Natural Science Foundation of China under Key Program Grant No. 62034007, 
the Science and Technology Innovation Key R\&D Program of Chongqing under Grant No. CSTB2023TIAD-STX0001, 
and the Hong Kong Theme-based Research Scheme (TRS) project T45-701/22-R of the Research Grants Council (RGC).

\newpage
\newpage
\bibliographystyle{IEEEtran} 
\bibliography{custom}

\end{document}